\documentclass[journal ]{new-aiaa}
\usepackage[utf8]{inputenc}
\usepackage{textcomp}

\usepackage{graphicx}
\usepackage{amsmath}
\usepackage[version=4]{mhchem}
\usepackage{siunitx}
\usepackage{longtable,tabularx}
\usepackage{amssymb}
\title{Deep Learning Enabled Uncorrelated Space Observation Association}

\author{Jacob J. Decoto \footnote{Non-Degree Option (NDO) Student, Computer Science, 450 Serra Mall Blvd, Stanford, CA, 94305}}
\affil{Stanford University, Stanford, CA, 94305, USA}

\author{David RC. Dayton\footnote{Software Engineer, Colorado Springs, CO 80831}}
\affil{Colorado Springs, CO, 80831, USA}

\begin{document}

\maketitle

\begin{abstract}
Uncorrelated optical space observation association represents a classic needle in a haystack problem.  The objective being to find small groups of observations that are likely of the same resident space objects (RSOs) from amongst the much larger population of all uncorrelated observations.  These observations being potentially widely disparate both temporally and with respect to the observing sensor position. By training on a large representative data set this paper shows that a deep learning enabled learned model with no encoded knowledge of physics or orbital mechanics can learn a model for identifying observations of common objects.  When presented with balanced input sets of $50\%$ matching observation pairs the learned model was able to correctly identify if the observation pairs were of the same RSO  $83.1\%$ of the time.  The resulting learned model is then used in conjunction with a search algorithm on an unbalanced demonstration set of 1,000 disparate simulated uncorrelated observations and is shown to be able to successfully identify true three observation sets representing 111 out of 142 objects in the population.  With most objects being identified in multiple three observation triplets.  This is accomplished while only exploring $0.06\%$ of the search space of 1.66e8 possible unique triplet combinations.
\end{abstract}

\section{Introduction}
\lettrine{S}{pace} Situational Awareness (SSA) relies heavily on optical sensors to provide angles only measurements of resident space objects (RSOs), especially in cases where radar observations would not be available.  Optical space surveillance sensors can range from networks of relatively cheap commercial sensors, to large aperture university or government telescopes, and even on orbit sensors which are unaffected by weather and can provide enhanced angular diversity in measurements.  From a high level view, the processing chain for the raw imagery from these sensors consists of pulling out detections of objects from the images, building tracks of detections over time, and correlating these tracks to known RSOs.  The remaining detection tracks that cannot be correlated to any known object with high confidence are known as Un-Correlated Object (UCO) detections or simply UCOs.  

UCOs can be caused by false positive detections, a known object for which the current catalog orbit state knowledge is poor such as with a recently maneuvered or high area to mass ratio object, or an unknown satellite or piece of debris that is not in the catalog.  Since a sensor typically takes a series of images in quick succession and stitches together tracks out of the detections, each UCO detection is a series of bearing angles over a typically short time duration.  The duration that a single UCO is tracked could be on the order of seconds or minutes, and in rarer cases even hours.  In the case that the UCO is tracked for only a short time, because of the short observed arc and angles only nature of the measurements, it is likely to be infeasible to determine the orbit of the unknown object with enough accuracy to re-task sensors to perform follow on collects.  Without determining the orbit, the object remains lost and will likely show up in subsequent collects as another UCO, exacerbating the problem of maintaining situational awareness.

\begin{figure}[t]
\begin{center}
\fbox{\includegraphics[width=14cm]{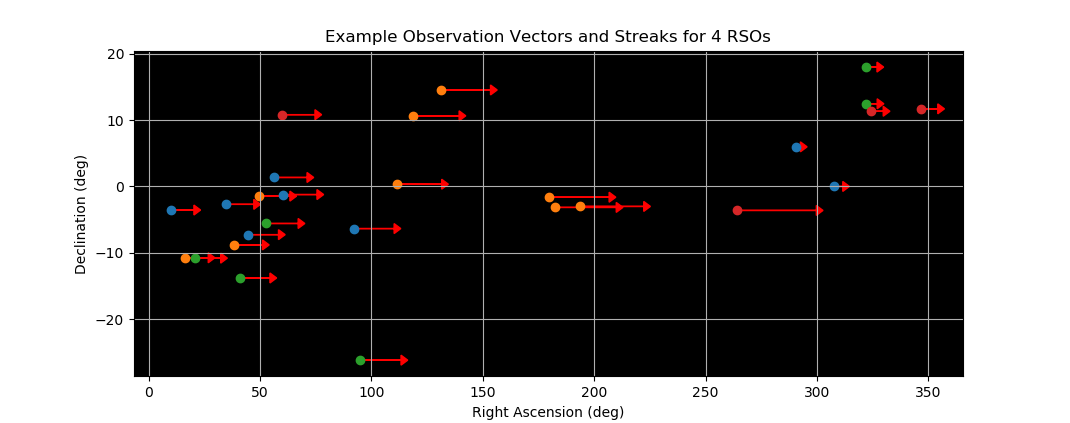}}
\end{center}
   \caption{Example of observation tracks of four different RSOs from different sensors and epochs projected onto a 2D plane}
\label{fig:images}
\end{figure}

The solution to this problem is to attempt association to one another of individual UCO observations from different images taken by various sensors usually at different times.  An illustration of this problem is shown in Figure \ref{fig:images} for a toy example with observations of four RSOs taken by different sensors at different times. Once multiple tracks, ideally from angularly disparate sensors with great temporal separation, are associated, a more accurate orbit can be fit to the unknown object than would be possible with only a single track.  If this orbit is accurate enough to allow for follow up collections, the UCO can be confirmed and the orbit can be refined further, leading to a new object, or recovered lost object, being maintained.  

This paper explores a novel data driven approach to UCO association which learns by example.  In this method, no knowledge of orbital mechanics or physics is encoded in the algorithm.  Instead a deeply layered neural network is trained on the data to make predictions about whether pairs of observations are of the same RSO or not.  The method is applied to a representative simulated data set with the goal of associating as many possible matches as possible while minimizing false positive associations which would have a cost in terms of compute time if they were passed to an Initial Orbit Determination (IOD) process. 

\section{Related Work}

Several published approaches to addressing the angles only observation association problem exist  \cite{Aristoff} \cite{Azimov} \cite{DeMars} \cite{Kelecy} \cite{Rutten} \cite{Stauch}.  Generally, these approaches use some form of analytically solved for Constrained Admissible Region (CAR), combined with application of algorithmic means of handling uncertainty and multiple association hypothesis.  Examples of methods employed include joint probabilistic data association (JPDA) and belief propagation (BP) \cite{Rutten}.   As described in Aristoff \cite{Aristoff}, none of these methods are a one sized fits all solution.  With some performing better in certain conditions than others.  While data driven machine learning techniques have been applied to orbit determination, \cite{Lee} \cite{Peng} \cite{Sharma}, to the author's knowledge, no published examples exist of applying purely data driven approaches to the uncorrelated observation association problem directly. 

\section{Dataset and Features}

Fortunately, the problem of space observation association allows for relatively easy simulation of large amounts of representative training data.  For this experiment completely separately generated training, validation, and test data sets were built.  With each consisting of 100,000 simulated ground sensor observations of a different population of satellites over a 12 hour period. Observed satellites were generated randomly from a uniform distribution within the ranges of orbital parameters outlined in Table \ref{table:satorbits}.  Each simulated satellite was observed by different randomly placed earth based sensors between 3-10 times at random epochs within the 12 hour window.  Each angles only observation was simulated as being observed over a period of 120 seconds. Additionally, random Gaussian noise, mean zero and standard deviation 100 meters, was modelled on the satellite position to introduce angular error in the measurements.  Each resulting observation consists of 10 parameters; epoch, observer position components, observer to RSO unit vector, and rate of change of observer to RSO unit vector.  

\begin{table}[ht]
\caption{Range of Uniform Distribution of Randomly Created Satellite States}
\centering 
\begin{tabular}{c c c c} 
\hline\hline
Orbital Parameter & \ Range & \ Units \\ [0.5ex] 
\hline
Semi-Major Axis & 41,164-43,164 & km \\ 
Eccentricity & 0.0-0.1 & \\
Inclination & 0-20 & deg \\
Longitude of Ascending Node & 0-360 & deg \\
Argument of Perigee & 0-360 & deg \\
Mean Anomaly & 0-360 & deg \\
\hline
\label{table:satorbits}
\end{tabular}
\end{table}

\section{Method}

\subsection{Data Preparation}

In order to give the neural network the best chance of learning a function that would generalize well, several operations were performed on the 10 base parameters in the bearings only observational data.  First, the observer position magnitude was added as an eleventh feature and the existing position vector converted to a unit vector.  The streak magnitude over time was then added as a twelfth parameter. Data points were then built from randomly selected pairs of observations.  For each pair a series of dimensional operations were performed where the difference in selected parameters from each observation in the pair were divided by the the difference in other selected parameters from the first observation in the pair.  Several variations of included parameters in the dimensional operations were tried.  The fundamental trade off being providing more potential features of use to the model versus slowing down training times and limiting the amount of data points processed in each batch due to memory limitations.

Equation \ref{eq1} gives the operation for adding derived parameters.  Where $P$ is the set of derived parameters added to the 24 base parameters for the pair.  And where $ o_{x,y}$ is the value of a base feature y for observation x.  In the chosen architecture the intervals  $q \in Q$ and $r \in R$ contain seven parameters each while $s \in S$ contains eight.  For a total of $7 \times 7 \times 8 + 24 = 416$ features for each data point. This number of features is approximately equivalent to a 20x20 pixel gray scale image in terms of the length of each input vector.  Which when compared to most computer vision problems, for which deeply layered neural networks are routinely applied, is of a relatively small scale.

\begin{equation}
\label{eq1}
    P = \frac{o_{i,q} - o_{i-1,r}}{o_{i,s} - o_{i-1,s}} for {\, q\in Q}, {\, r\in R}, {\, s\in S}
\end{equation}

\subsection{Neural Network Classifier}

The neural network classifier \cite{GoodBengCour16} makes use of the PyTorch framework \cite{PyTorch} and consists of a a fully connected input layer, six fully connected hidden layers, and an output layer.  ReLU activation functions are used in all but the last layer and batch normalization is performed after each ReLU activation.  The final output layer uses a log softmax activation with two outputs corresponding to the probability that the two input observations are of the same RSO or two different RSOs.  For training purposes, randomly selected observation pairs were used, with an equal number of pairs consisting of two observations of the same RSO as those that did not. In the end application there would be many more pairs that were not of the same satellite, however the balancing is necessary in training to avoid learning a function which simply always predicts not a match.  

\begin{table}[hbt]

\caption{Train and Test Accuracy}
\centering 
\begin{tabular}{c c c c} 
\hline\hline
Data Architecture & \ Train Accuracy & \ Test Accuracy  \\ [0.5ex] 
\hline
24 Parameters & 60.3\% & 49.9\% \\ 
416 Parameters & 84.3\% & 83.1\% \\ 
600 Parameters & 90.2\% & 60.5\%  \\
\hline
\label{table:accuracy}
\end{tabular}
\end{table}

 For training, a standard Adam Optimizer was used with learning rate varied between 1e-4 to 1e-6. Cross-entropy loss is used as the loss function.  Training was performed with sets of 40,000 data points with evaluation on a validation and test set of 4,000 points each.  Table \ref{table:accuracy} shows the best achieved test accuracy and corresponding training accuracy for three different derived feature architectures.  Without performing the dimensional derived data augmentation described previously the resulting 24 parameter data structure resulted in no significant learning.  However, by augmenting the data as described, the 416 parameter architecture was able to achieve a test set accuracy  of $83.1\%$ while keeping over fitting to a minimum as evidenced by the only slightly high training set accuracy.  All further presented results utilize the learned model from the 416 parameter variation.

\begin{figure}[hbt]
\begin{center}
\fbox{\includegraphics[width=14cm]{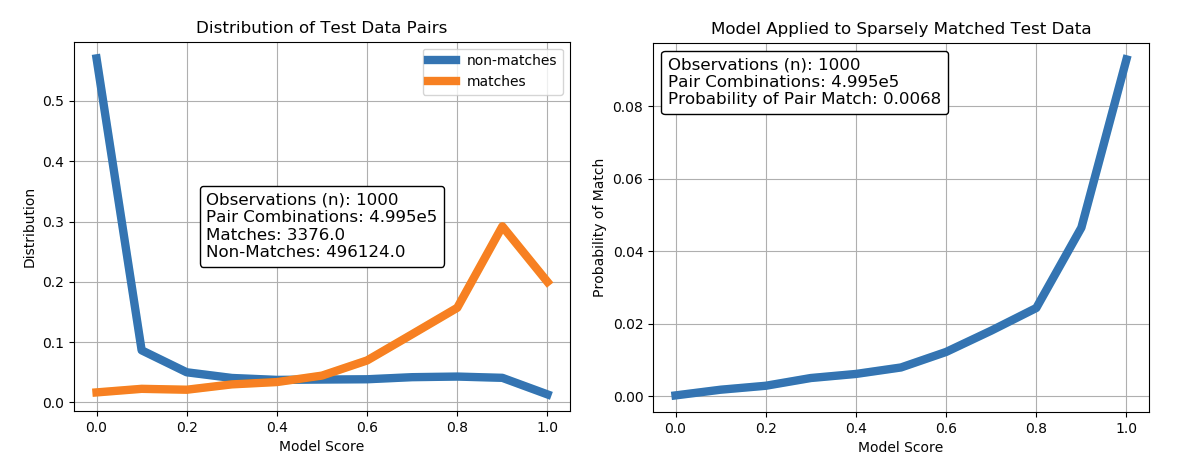}}
\end{center}
   \caption{Distribution of matching and non-matching pairs from test set (Left) and probability of a match given model score (Right)}
\label{fig:model_stats}
\end{figure}

Post training analysis used a 1,000 observation sub-set of the test data.  In this subset a total of $4.9e5$ unique observation pairs are possible with only 3,376 of the possible pairs containing true matches.  Figure \ref{fig:model_stats} at left shows the learned model probability of a match score for all $4.9e5$ combinations when the true pair was a match versus not a match.  The vast majority of non-matching pairs received scores of less than 0.2 while the true matches were heavily weighted towards the upper end of the probability score.  Figure \ref{fig:model_stats} at right bins the $4.9e5$ datapoints by score and reports the percent of datapoints that were a match by score.  At the high end of the scale, points receiving a score greater than 0.95 had a true probability of being a match of greater than 0.09 while the overall probability in the data set was 0.0068.  This illustrates that while nothing is certain, the learned model can provide a much greater chance of associating observations than brute force guessing.

\subsection{Application of Learned Model}

The end goal of an observation association algorithm is to correctly identify groups of enough observations of the same RSO to enable IOD with sufficient accuracy to gain custody of the object.  For purpose of the following experiment, it is assumed that a chain of three observations from different sensors and different times is sufficient.  A sub-set of the test data consisting of 1,000 observations of 147 different unique RSOs was used to test application of the learned model for predicting probability of a match.  Starting with each of the 1,000 observations in sequence, the pre-trained neural network was used to obtain a match score of a pairing of the base observation with each of the others.  Uniform Cost Search was then used to identify the top N candidate three observation triplets which contained the base observation.   

\begin{figure}[t]
\begin{center}
\fbox{\includegraphics[width=12cm]{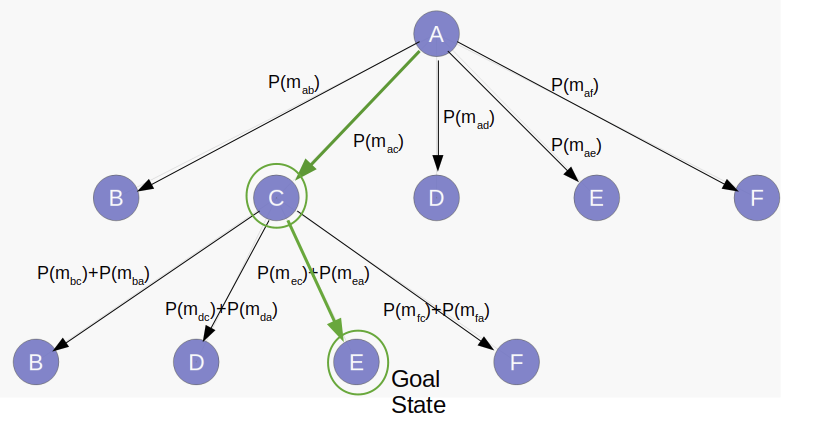}}
\end{center}
   \caption{Illustration of Uniform Cost Search tree for toy example}
\label{fig:search_tree}
\end{figure}

An example of this search process is shown in Figure \ref{fig:search_tree} for a six observation example.  In this example the base observation is Node A.  The first node to be expanded is that with the lowest cost from Node A.  The cost is equal to the probability from the neural network model that the observation represented by $Node_i$ is not of the same RSO as $Node_j$.  In the illustrated example the lowest cost, highest probability of a match, node to expand is Node C.  When expanding nodes to the next depth layer the added cost is a function of the probability of no match with the node at the deepest layer with all nodes up the chain.  For example the cost of expanding to Node E is the sum of the probability of no match between Node E and Node C and also that of Node E and Node A.

\begin{figure}[ht]
\begin{center}
\fbox{\includegraphics[width=8cm]{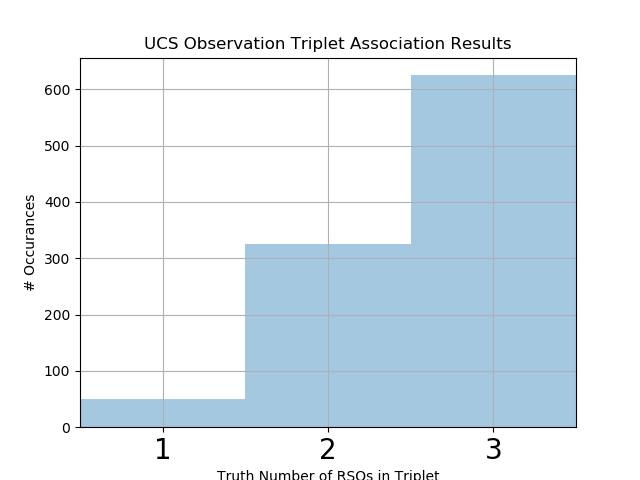}}
\end{center}
   \caption{Number of unique RSOs in each of 1,000 explored triplets (1 = All Matching)}
\label{fig:triplet_stats}
\end{figure}

In order to reduce the branching factor and speed up search an imperfect heuristic was used that eliminated nodes for which the probability score of no match with the base Node was above a threshold d, set to 0.3. Also variable was the number of solutions, s, for the algorithm to return for each base Node.  The number of total candidate triplets identified, that would be sent to an IOD algorithm, is equal to s times the number of observations.  Figure \ref{fig:triplet_stats} shows the truth number of RSOs represented in each triplet when $s=1$.  When taking the top candidate only, $s=1$, 50 of the 1,000 candidates triplets contained observations all of the same RSO.  These 50, under the previously stated assumption, would be likely to result in an orbit fit of sufficient accuracy to correlate further observations, refine the orbit of the RSO, and begin maintenance of the object in a catalog.  The other 950 triplet candidates would have resulted in either IOD failures, highly improbably orbit states, or plausible orbits but for which additional observations would be unlikely to correlate.  

\begin{figure}[ht]
\begin{center}
\fbox{\includegraphics[width=16.1cm]{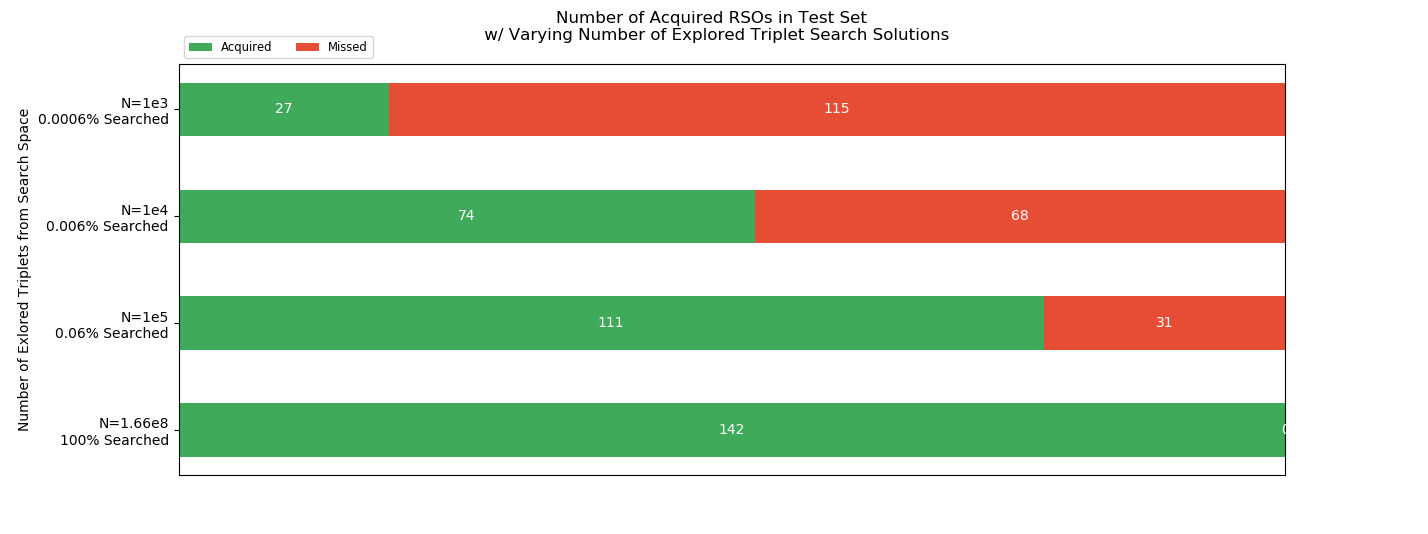}}
\end{center}
   \caption{Results of RSO acquisition experiment with 1,000 observation subset of test data}
\label{fig:summary_plot}
\end{figure}

In the case of the test data set, the true number of unique RSOs represented in the observations is 147.  Figure \ref{fig:summary_plot} shows the number of these RSOs for which a true triplet of observations was identified with varying settings of how many candidate solutions to return from the UCS algorithm.  The total number of possible solutions is given by Equation \ref{eq2} where n is the number of observations and r is the desired number of observations to group.  For 1,000 observations with triplets as the desired output there are $1.66e^8$ unique combinations. Figure \ref{fig:summary_plot} shows that by utilizing the pre-trained neural network model to predict probabilities of match and the UCS algorithm to identify most likely triplets, that only $0.06\%$ of the search space need be sent to an IOD algorithm to acquire orbits on 111 out of 147 of the RSOs in the example test set.

The utilized approach assumed that all candidates would be identified prior to sending candidates to an IOD algorithm.  In actual practice it may be desirable to send top candidates early and use the solved for states to correlate further observations.  In this way, the candidate pool of observations could be reduced as the number of UCOs becomes smaller, simplifying the problem and reducing the false positive rate of subsequent batches of candidate groupings.

\begin{equation}
\label{eq2}
    Unique Combinations = f(n,r) = \frac{n!}{(n-r)!r!}
\end{equation}

\section{Saliency Maps}

In order to examine what features the learned model was using to make predictions about whether observations were of the same RSO, a series of saliency maps were constructed.  The 416 element feature vector for pairs of observations was transformed into a 21x21 pixel grayscale image, with null values in the lower right to fill out unused pixels.  Next a probability score corresponding to 1.0 for the true class, match or no match, was back propagated through the neural network and the resulting gradients over the input pixel values recorded.  In this way the saliency map can show for a given data point, represented here by an image, which of the features contributed most to the classification.  Examples of saliency maps for randomly selected non-matching and matching pairs are shown in Figure \ref{fig:saliency_examples} of the Appendix.  A high degree of variability was observed in what features were contributing most to the classification.  Indicating that a few strong features were not dominating the classification.

To determine if certain features were contributing more heavily on average, a composite saliency map was built for 1,000 samples each of non-matching and matching data points, shown in Figure \ref{fig:saliency_composite} of the Appendix.  Certain features did play more prominently, however they appeared to be different for matching versus non-matching data points.

\section{Conclusions and Further Work}

This paper demonstrates a data driven approach to solving the observation association problem.  This approach uses no encoded knowledge of physics or orbit mechanics, and for the example problem was able to identify correct observation triplet associations for $76\%$ of RSOs in a sample data set while only exploring $0.06\%$ of the possible combinations.  It should be noted, that while the model training was computationally expensive, application of the learned model is very fast to process, with on the order of 1e6 observation pairs processed on a desktop class machine in under a second.  A logical next step would be to apply this approach to a real world data set and if possible compare performance with known truth data to current expert systems.  While it seems unlikely that this data driven learned model would replace expert systems, the data driven approach could be used to augment and could provide several key advantages.  Foremost, it is amenable to including additional features that would be difficult to account for in current systems.  For example, observations typically include a measurement of detection brightness relative to the starfield background.  With this approach, brightness or other signature characteristics could be readily incorporated in the data features and the model re-trained.  Other advantages include the the ability to train models to be better at detecting specific classes of objects.  If a large data set were available of observations of known high area to mass ratio (HAMR) objects, for example, then a specialized model could be trained with this data set that may outperform other methods for this specific subset of objects.  Finally, the learned approach is adaptable as the data characteristics and makeup change over time, without need for human intervention other than to retrain the model.  It is also possible that a combined data driven and physics based expert system approach would be desirable.  Using a constrained admissible region approach for example, additional strong features could be added to the data points, potentially boosting the performance of the model.

\bibliography{sample}

\clearpage

\section{Appendix}

\begin{figure}[hbt]
\begin{center}
\fbox{\includegraphics[width=16cm]{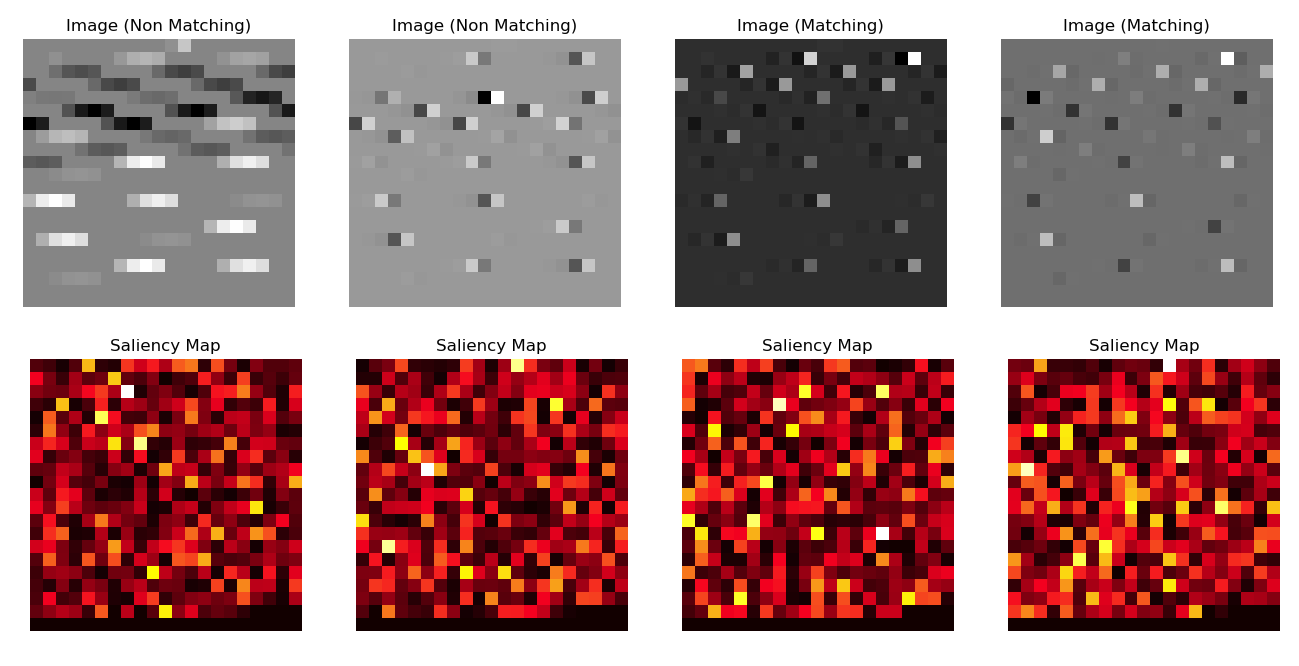}}
\end{center}
   \caption{Image representations of data examples and corresponding saliency maps}
\label{fig:saliency_examples}
\end{figure}

\begin{figure}[thb]
\begin{center}
\fbox{\includegraphics[width=10cm]{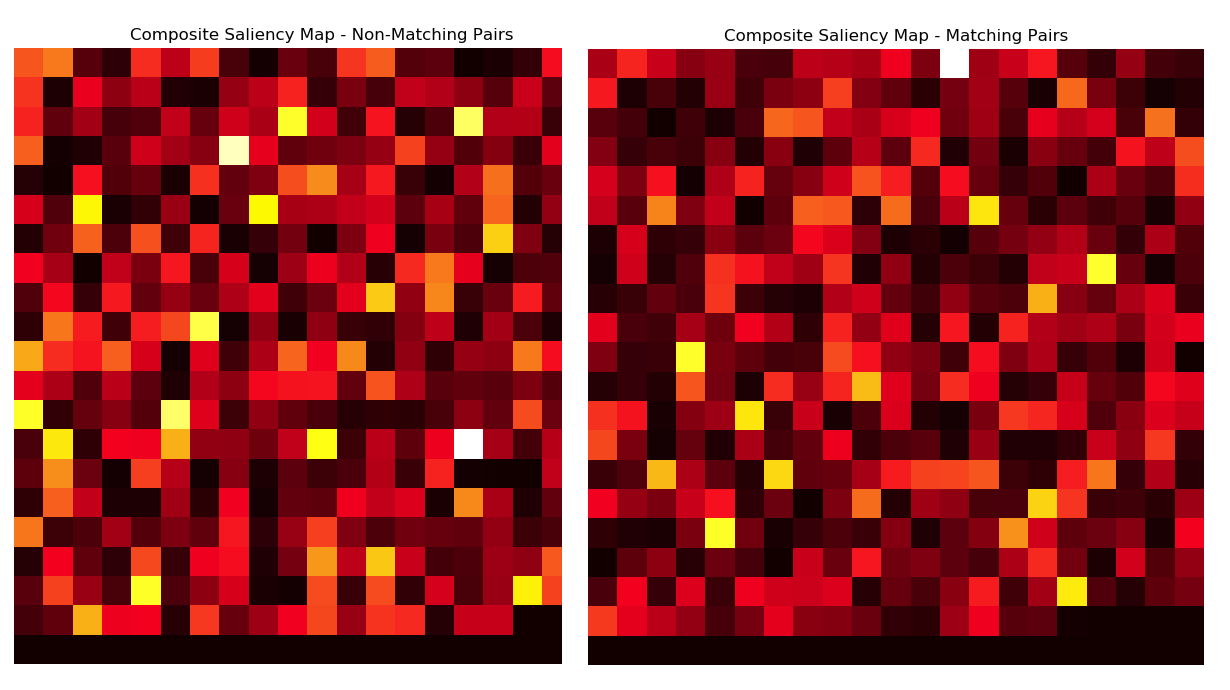}}
\end{center}
   \caption{Composite saliency maps for 1,000 non-matching (Left) and matching (Right) datapoints}
\label{fig:saliency_composite}
\end{figure}

\end{document}